\documentclass[11pt,a4paper]{article}
\usepackage[hyperref]{acl2019}
\usepackage{times}
\usepackage{latexsym}
\usepackage{graphicx}
\usepackage{framed}
\usepackage{algorithm}
\usepackage{tikz}
\usetikzlibrary{automata, positioning, arrows}
\usepackage{url}

\aclfinalcopy 
\def\aclpaperid{5} 

\tikzset{->, >=stealth', node distance=1.5cm, every state/.style={thick, fill=gray!10}, font={\fontsize{8pt}{6}\selectfont}}

\newcommand{\be}{\begin{enumerate}}
\newcommand{\ee}{\end{enumerate}}
\newcommand{\bi}{\begin{itemize}}
\newcommand{\ei}{\end{itemize}}
\newcommand{\bit}{\begin{itemize}}
\newcommand{\eit}{\end{itemize}}
\newcommand{\ben}{\begin{enumerate}}
\newcommand{\een}{\end{enumerate}}

\newcommand{\temv}[1]{\mbox{\em #1\/}}

\newcommand{\etc}{\mbox{etc.}}

\newcommand{\egnsp}{\mbox{e.g.}}

\newenvironment{solutionFollowedByNewParagraph}{{\bf Solution}. }{\vspace*{-0.03in} \\}

\newcommand{\bs}{\begin{solution}}
\newcommand{\es}{\end{solution}}
\newcommand{\bsnp}{\begin{solutionFollowedByNewParagraph}}
\newcommand{\esnp}{\end{solutionFollowedByNewParagraph}}

\newcommand{\nth}[1]{\mbox{$#1^{\mbox{\em th}}$}}

\newcommand{\card}[1]{\mbox{$|#1|$}}

\newcommand{\mlb}{\textnormal{[\kern-.15em[}}
\newcommand{\mrb}{\textnormal{]\kern-.15em]}}

\newcommand{\msp}{\mbox{$\;\:$}}

\newcommand{\bcen}{\begin{center}}
\newcommand{\ecen}{\end{center}}

\newcommand{\aaatran}[1]{\text{$\mathfrak{A}_3$}}

\newcommand{\KA}[1]{K.A. Comment: {\color{red} #1}}
\renewcommand{\KA}[1]{}

\newcommand{\mainmachinesp}{$M_{\mbox{\em\scriptsize LF}}$ }  

\title{Grammatical Sequence Prediction for \\ Real-Time Neural Semantic Parsing}

 \author{Chunyang Xiao \\
  Bloomberg \\ London \\ United Kingdom  \\
  \texttt{\small cxiao35@bloomberg.net} \\\And
  Christoph Teichmann  \\
  Bloomberg \\ London \\ United Kingdom \\
  \texttt{\small cteichmann1@bloomberg.net} \\\And  
  Konstantine Arkoudas \\ 
  Bloomberg \\ New York \\ USA \\
  \texttt{\small karkoudas@bloomberg.net} \\}

\date{}

\begin{document}
\maketitle
\begin{abstract}

While sequence-to-sequence (seq2seq) models achieve state-of-the-art performance in many natural language processing tasks,
they can be too slow for real-time applications. One performance bottleneck is predicting the most likely next token over a large vocabulary;
methods to circumvent this bottleneck are a current research topic. We focus specifically on using seq2seq models for
semantic parsing, where we observe that grammars often exist which specify valid formal representations of utterance semantics.
By developing a generic approach for restricting the predictions of a seq2seq model to grammatically permissible continuations,
we arrive at a widely applicable technique for speeding up semantic parsing. The technique leads to a 74\% speed-up 
on an in-house dataset with a large vocabulary, compared to the same neural model without grammatical restrictions. 

\end{abstract}

\section{Introduction} Executable semantic parsing is the task of mapping an utterance
to a logical form (LF) that can be executed against a data store (such as a SQL database or a knowledge graph),
or interpreted by a computer program in some other way.\footnote{From here on we will refer to executable semantic
  parsing simply as semantic parsing.} Various authors have tackled this task via
sequence-to-sequence (seq2seq) models, which have already led to substantial advances in machine translation.
These models learn to directly map the input utterance into a linearised representation
of the corresponding LF, predicting it token by token. Seq2seq approaches have yielded
state-of-the-art accuracy on both classic (e.g., Geoquery \cite{Zelle:1996:LPD:1864519.1864543} and Atis \cite{Dahl:1994:ESA:1075812.1075823}) and more recent semantic parsing
datasets (e.g., WebQuestions, WikiSQL and Spider)~\cite{DBLP:journals/corr/LiangBLFL16,DBLP:conf/acl/DongL16,DBLP:journals/corr/abs-1805-04793,DBLP:journals/corr/abs-1810-02720, DBLP:journals/corr/abs-1810-05237}. The recent datasets are of much larger scale,
which not only enables the use of more data-hungry models, such as deep neural networks,
but also provides more complex challenges for semantic parsing.

The material presented in this paper was motivated by a question-answering dataset
for equity search in a financial data and analytics system. 
We will refer to this dataset as ``the EQS dataset'' going forward (and we will refer to ``equity search'' as EQS for short).
The queries in the dataset pertain to equity stocks; they are usually of the form
{\em Show me companies that satisfy such-and-such criteria\/}, or {\em What are the top 10 companies that $\cdots$?},
and so on. The dataset pairs such queries with logical forms that capture their semantics. These
logical forms are designed to be readily translatable into an executable query language in order to retrieve
the corresponding answers from a data store in the back end. 
Questions can involve a large number of diverse search criteria, such as price,
earnings per share, country of domicile, membership in indices, trading in specific exchanges,
etc., applied to a large set of equities for which the system offers information.

The large number of search criteria
and entities is reflected in the LFs, leading to a problem common with newer, more complex
semantic-parsing datasets: having to deal with a large LF vocabulary size. In the EQS dataset the LF
vocabulary has a size that exceeds 50,000. Since seq2seq models apply some operation over the whole vocabulary -- usually
the softmax operation -- when deciding what symbol to output next, large LF vocabularies can 
slow them down considerably. For example, we observe in our EQS experiments with seq2seq models
that it takes on average between 250 and 300 milliseconds to parse a query,
which is too slow for one single component in a larger, real-time question-answering pipeline.
This is consistent with observations made previously in the neural language modelling
literature; see for example \newcite{Bengio:2003:NPL:944919.944966, conf/interspeech/MikolovKBCK10}, 
where the authors show that when the vocabulary size exceeds a certain threshold,
the softmax calculation becomes the computational bottleneck. 

Our proposal for tackling this bottleneck is based on the fact that there
generally exist grammars, which we call {\em LF grammars},
specifying the concrete syntax of valid logical forms (LFs). This is usually the case
because LFs need to be machine-readable. We further note that,
for a given LF prefix, one can usually use the LF grammar to look up
the next grammatically permissible tokens (i.e., tokens that are part of a
grammatically valid completion of the prefix). For example, if the language of valid LFs
can be expressed by a context-free grammar (CFG), as is almost always the case,
then look-ups could be performed with an online version
of the Earley parser~\cite{Earley70}. If it is possible to efficiently 
look up the permissible next tokens for a given prefix, then
restricting the softmax operation to those permissible tokens should improve efficiency, and
because only non-permissible tokens are ruled out, this will only ever prevent the system from
producing invalid LFs.

If the number of grammatically permissible tokens at some prediction step 
is substantially smaller than the LF's vocabulary size, the integration
of the LF grammar may reduce prediction time for that step significantly.
In semantic parsing problems a grammar can naturally lead to prediction
steps with few choices. To see why this might be the case, consider our LFs in Figure~\ref{fig:LF_examples}, 
which involve atomic constraints of the form:
$$(\mbox{\em field\/} \;\: \mbox{\em operator} \;\: \mbox{\em value\/}).$$
While there are many grammatically permissible choices for \mbox{\em field\/} and
\mbox{\em value\/}, the
choices for \mbox{\em operator\/} are rather limited.\footnote{Equality, less than and so on.} LFs for many applications
will contain ``structural'' elements with a limited number of choices in grammatically predictable positions,
and we can use grammars to exploit this fact.

In order to make the computation of permissible next tokens efficient,
we propose to use a finite-state automaton (FSA) approximation of the LF grammar. Finite-state automata
can capture local relations that are often quite predictive of the admissible tokens in a given context, 
and can therefore lead to considerable speed improvements for our setting, even if we use an approximate
grammar. Moreover, approximations can be
designed in such a way that a FSA accepts a superset of the actual LF language,
preserving the guarantee that only ill-formed LFs will ever be ruled out.
 

In this paper we therefore work with a grammar for which the next
permissible tokens can be computed efficiently, and show how such a grammar
can be combined with a seq2seq model in order to substantially improve the
efficiency of inference. While we focus on using
FSAs to restrict a recurrent neural network with
attention in the EQS dataset, our approach is generic and could be used
to speed up any sequential prediction model with any grammar that allows
for efficient computation of next-token sets. Our experiments show that in 
our domain of interest we obtain a reduction in parsing time by up to 74\%.

\section{Logical Forms and their Grammar}
\begin{figure}[t]
	\begin{center}
	\noindent{%
		\parbox{\linewidth}{%
			\textbf{Query:} return on capital sp$500$
			
			\textbf{LF:} \begin{scriptsize}
				\textit{(AND}\\
				\textit{\hspace*{10mm}(FLD\_INDEX EQ enumValue(IDX\_SP500))}\\
				\textit{\hspace*{10mm}(display FLD\_RETURN\_ON\_CAP))}
			\end{scriptsize}
			
			\textbf{Query:} steel western europe not german
			
			\textbf{LF:}\begin{scriptsize}
				\textit{(AND}\\
				\textit{\hspace*{10mm}(NOT (FLD\_DOMICILE EQ enumValue(COU\_GERMANY)))}\\
				\textit{\hspace*{10mm}(FLD\_DOMICILE EQ enumValue(COU\_WESTERN\_EUROPE))}\\
				\textit{\hspace*{10mm}(FLD\_EQS\_SECTOR EQ enumValue(SEC\_GICS\_STEEL)))}
			\end{scriptsize}
		}%
	}
\end{center}
	\caption{Two (query, LF) pairs in the EQS dataset.}\label{fig:LF_examples}
\end{figure}
\subsection{Equity search}

The domain of interest is that of {\em equity search}, or {\em EQS\/} for short, in which queries
are intended to screen for companies\footnote{Or more precisely, for tradeable equity {\em tickers\/} such
  as {\tt IBM} or {\tt FB}.} that satisfy certain criteria, such as being domiciled in a certain
country or region (such as France or North America), being in a certain sector (such as the automobile
or technology sectors), being members of a certain index (such as the S\&P 500), being traded
in certain exchanges (such as the London or Oslo stock exchanges), or their fundamental financial
indicators (such as market capitalization or earnings per share) satisfying certain simple
numeric criteria. Some sample queries:
\begin{itemize}
\item {\em What are the top five Asian tech companies?}
\item  {\em Show me all auto firms traded in Nysex whose market cap last quarter was over \$1 billion}
\item {\em Top 10 European non-German tech firms sorted by p/b ratio}
\end{itemize}
Queries may also be expressed in much more telegraphic style, e.g., the second query
could also be phrased as {\em auto nysex last quarter mcap $>$ \$1bn}. The two queries
in Figure~\ref{fig:LF_examples} are additional examples of tersely formulated queries,
the first one asking to display the return-on-capital for all companies in the S\&P 500
index, and the second one asking for all Western European companies in the steel sector
except for German companies. 

The LF language we use was designed to express the search intent of a query
in a clear and non-ambiguous way. In the following section we describe the
abstract grammar and concrete syntax of a subset of this LF (we cannot treat
every construct due to space limitations). 

 
\subsection{LF Abstract Grammar and Concrete Syntax}
As with many formal logical languages, the abstract grammar of our LF naturally
falls into two classes: {\em atomic\/} LFs corresponding to individual logical
or operational constraints; and {\em complex\/} LFs that contain other LFs as
proper parts. The former constitute the basis case of the inductive definition
of the LF grammar, while the latter correspond to the recursive clauses.

\paragraph{Relational Atomic Constraints}

The main atomic constraints of interest in this domain are relational, of the form
$$(\temv{field}(\temv{t}) \msp\temv{op} \msp\temv{value})$$
where typically \temv{field} is either a numeric field (such as \temv{price});
or a so-called ``enum field,'' that is, an enumerated type.
An example here would be a field such as a credit rating (say, long-term Fitch ratings),
which has a a finite number of values (such as \temv{B+}, \temv{AAA}, \etc); 
or country of domicile,
which also has a a finite number of values (\temv{algeria}, \temv{belgium}, and so on);
or an index field, whose values are the major stock indices (such as the S\&P 500). 
The \temv{value} is a numeric value if the corresponding \temv{field} is numeric,
though it may be a {\em complex numeric value}, \egnsp, one that has currencies or denominations
attached to it (such as ``5 billion dollars''). The operator \temv{op} is either equality (\temv{EQ}),
inequality (\temv{NEQ}), less-than (\temv{LS}), greater-than (\temv{GR}), less-than-or-equal (\temv{LE}),
etc.\footnote{If the field is an enum, then comparison operators such as \temv{GR} or \temv{LE} make
  sense only if the field is ordered. Credit ratings are naturally ordered, but countries, for example, are not.
  Nevertheless, the {\em syntax\/} of constraints allows for $(\temv{france}\msp GE \msp 2)$; such a constraint is weeded out
  by {\em type judgments}, not by the LF grammar.}
  Note that all fields, both numeric and enum, 
are indexed by a time expression $t$, representing the value of that field at that particular
time. For example, the atomic constraint $$(\temv{price}(\temv{June} \msp 23, 2018) = \$100)$$
states that the (closing) price on June 23, 2018 was 100 USD. We drop the time $t$ when
it is either immaterial or the respective field is not time sensitive. We omit the specification
of the grammar and semantics of time expressions, since we will not be using times in what follows
in order to simplify the discussion.

\paragraph{Display Atomic Constraints}

Some of our atomic constraints are operational in the sense that they
represent directives about what fields to {\em display\/} as the query
result, possibly along with auxiliary presentation information such as sorting order.
For instance, for the query {\em Show me the market caps and revenues of asian tech firms},
two of the resulting constraints would be the display directives 
$(\temv{display} \msp \temv{FLD\_MKT\_CAP})$ and
$(\temv{display} \msp \temv{FLD\_SALES\_REV\_TURN})$. 

\paragraph{Complex Constraints}

Complex constraints are boolean combinations of other constraints, obtained by applying one of the
operations \textit{NOT, OR, AND}, resulting in recursively built constraints of the form
$(\temv{NOT}\msp c)$, $(\temv{AND}\msp c_1 \msp c_2 \msp \cdots \msp c_n)$, and $(\temv{OR}\msp c_1 \msp c_2)$.\footnote{We model the $OR$ operation as binary operation and the $AND$ operation as n-ary to make them close to the natural language syntax we observe in the dataset.} 
 

\section{Encoding LF grammar in FSAs}\label{sec::aut}

\begin{figure*}[t]
\tikzset{initial text = $RCM$,}
\begin{tikzpicture}
\node[state, initial] (q0) {$0$};
\node[state, right of=q0] (q1) {$1$};
\node[state, above right of=q1, xshift=2cm, yshift=1cm] (q2) {$2$};
\node[state, right of=q2] (q3) {$3$};
\node[state, below right of=q1, xshift=2cm, yshift=-1cm] (q4) {$4$};
\node[state, right of=q4] (q5) {$5$};
\node[state, below right of=q3, xshift=2cm, yshift=-1cm] (q8) {$8$};
\node[state, accepting, right of=q8] (q9) {$9$};
\node[state, right of = q1, xshift=2cm](q6) {$6$};
\node[state, right of = q6](q7) {$7$};
\draw (q0) edge[above] node{(} (q1)
(q1) edge[above] node[align=left]{\begin{minipage}{1.2in} FLD\_DOMICILE, \\ FLD\_INDEX, \\ $\cdots$ \end{minipage}}(q2)
(q2) edge[above] node[align=left]{EQ}(q3)
(q3) edge[above] node{\hspace*{1.2in}\begin{minipage}{1.2in} COU\_GERMANY, \\ IDX\_SP500, \\ $\cdots$ \end{minipage}}(q8)
(q1) edge[below] node[align=left]{\begin{minipage}{1.2in}FLD\_EPS, \\ FLD\_PE\_RATIO, \\ $\cdots$\end{minipage}}(q4)
(q4) edge[below] node[align=left]{EQ, \\ LT \\ $\cdots$ \vspace*{0.2in}}(q5)
(q5) edge[below] node{\hspace*{0.3in}\temv{FPNM}}(q8)
(q1) edge[above] node{\hspace*{0.78in}\begin{minipage}{1.2in} FLD\_MOODY, \\ FLD\_FITCH, \\ $\cdots$ \end{minipage}}(q6)
(q6) edge[above] node[align=left]{EQ, \\ LT \\ $\cdots$}(q7)
(q7) edge[above] node{\hspace*{0.8in}\begin{minipage}{1.2in} AA,\\ B-, \\ $\cdots$ \end{minipage}}(q8)
(q8) edge[above] node{)}(q9);
\end{tikzpicture}

\tikzset{initial text = $OR\;Machine$,}
\begin{tikzpicture}
\node[state, initial] (q0) {$0$};
\node[state, right of = q0] (q1) {$1$};
\node[state, right of = q1] (q2) {$2$};
\node[state, right of = q2, xshift = 1cm] (q3) {$3$};
\node[state, right of = q3, xshift = 1cm] (q4) {$4$};
\node[state, accepting, right of = q4] (q5) {$5$};
\node[state, below of = q2] (q6) {$6$};
\node[state, accepting, below of = q6] (q7) {$7$};
\node[state, below of = q3] (q8) {$8$};
\node[state, accepting, below of = q8] (q9) {$9$};
\draw (q0) edge[above] node{(} (q1)
         (q1) edge[above] node{OR} (q2)
         (q2) edge[above] node{RCM} (q3)
         (q3) edge[above] node{RCM} (q4)
         (q4) edge[above] node[align=left]{)} (q5)
         (q2) edge[left] node[align=left]{(} (q6)
         (q6) edge[left] node{\mbox{OR,AND,NOT}} (q7)
         (q7) edge[loop left] node{RCM $\mid$ (,),OR,AND,NOT} (q7)
         (q3) edge[left] node[align=left]{(} (q8)
         (q8) edge[right] node{\mbox{OR,AND,NOT}} (q9)
         (q9) edge[loop right] node{RCM $\mid$ (,),OR,AND,NOT} (q9);
\end{tikzpicture}
	\caption{Some automata involved in building the supersert of the LF grammar.}\label{fig:automata}
\end{figure*}
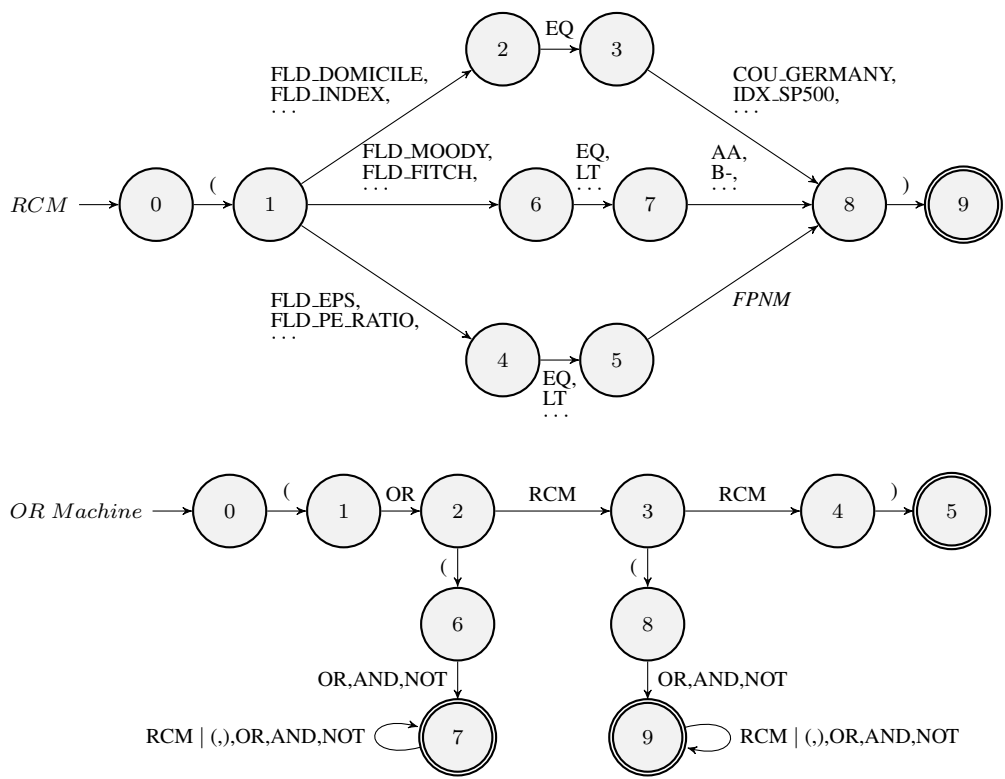



For efficient incremental parsing and computation of the next permissible tokens, we encode our grammar
using finite state automata (FSAs). As FSAs can only produce regular languages that are strictly less
expressive than context free languages such as the one recognized by our LF grammar,
our strategy is to use automata to build a superset for our LF language. Some of the automata
involved in building this superset are shown in Figure~\ref{fig:automata}. Note that, while we defined
our FSA approximation manually, there exist general techniques to construct
an automaton whose language is a superset of a CFG's language for any given CFG~\cite{Nederhof:2000:PER:971841.971843}.
This means that the approach could easily be used for any LF language that can be described by a CFG.

For all automata, we take the start states to be 0 and indicate the final states with double
circles. The ``$\mid$'' stands for the union operation over automata.
On each arc, we either specify as labels LF tokens that the FSA can consume in order
to transition to its next state(s); or else we specify a previously defined machine (automaton)
noted with ``$M$:\temv{machine\_name}''\footnote{In that case the ``arc'' is just a concise representation of
  the entire automaton that goes by \temv{machine\_name}.}
where the source state of the arc coincides with the start state of the automaton 
and its target state coincides with the final state(s) of the automaton.\footnote
{In the case of multiple final states, one simply replicates the target state to coincide with each of the final states.} 
\KA{Generally speaking there can be many final states}

\paragraph{Relational Atomic Constraint machines}

The automaton \temv{RCM} (``Relational Constraint Machine'')
in the top part of  Figure~\ref{fig:automata} generates relational atomic
constraints of the form $(\temv{field}\msp\temv{op}\msp{value})$; the \temv{FPNM} (
floating point number machine) is an automaton recognizing restricted floating point numbers. 
Note that some extra-syntactic information about fields is explicitly built into the machine.
For example, if a constraint begins with an unordered enum field that only admits equality,
such as \textit{FLD\_DOMICILE}, then the operator (on the arc from state 2 to state 3)
is always \textit{EQ}, whereas if the field is ordered (as all numeric fields are, and some
enum fields such as ratings), then any operator may follow (such as \temv{LE}, \temv{GR}, \etc). 
The automaton constrains what follows a num field in a similar fashion. 

\paragraph{Complex Constraint machines}

Unlike their atomic counterparts, logically complex constraints
can be arbitrarily nested, thereby forming a non-regular context-free language 
that cannot be characterized by FSAs. We get around this limitation by
constructing FSAs for such complex constraints that accept a regular language
forming a {\em superset\/} of the proper context-free LF language.   The automaton ``OR Machine''
in Figure~\ref{fig:automata} illustrates such a construction. 
This machine recognizes LFs of the form $(\temv{OR\msp RCM\msp RCM})$
along the topmost horizontal path of the automaton (state sequence 0-1-2-3-4-5). 
But if one or two of these relational constraints are replaced by logically complex constraints,
the automaton can recognize the result by taking one or two of the vertical paths (state sequences 2-6-7 and
3-8-9, respectively). These paths can also accept strings that are not syntactically valid LFs.
However, we are only using these automata to restrict the softmax application  to a subset of the
LF vocabulary, and for that purpose these automata are conservative approximations. 
An alternative approach would be to use FSAs for logically complex constraints that
essentially unroll nested applications of logical operators up to some fixed depth $k$,
\egnsp, say $k = 2$ or 3, as logically complex constraints with more than 3 nested logical
operations are exceedingly uncommon, though possible in principle.
But the present approach is
simple and already leads to considerable reductions in the number of permissible tokens
at each prediction step, thereby significantly accelerating our neural semantic parser. 

The final automaton representing the entire LF language, which we write as $M_{\temv{\scriptsize LF}}$,
is the union of atomic machines such as \temv{RCM} with three ``approximation'' machines
for the three logical operators (negation, conjunction and disjunction).

\section{Combining Grammar and Neural Model}

\subsection{Grammatical continuations by Automata}

We now show how to use the automaton \mainmachinesp that represents the LF grammar
in order to (a) compute the set of valid next tokens, and (b) update the current
prefix by appending an RNN-predicted token. We present very simple algorithms for both operations, 
\textit{nextTokens} and \textit{passToken}, which can be used with any grammar that is represented as a
DFA.\footnote{For convenience, of course, the grammar could be represented by non-deterministic
  automata (NFAs). The algorithms we present here would still be applicable via 
  a simple preprocessing step that would convert the NFAs to DFAs using standard
  algorithms for that purpose \cite{Rabin:1959:FAD:1661907.1661909}.}

\textbf{\textit{nextTokens}}: This function returns a list of the permissible next tokens
based on the current automaton state, which corresponds to the current LF prefix
(note that because the grammar is a DFA, there is a unique resulting state  for
any prefix accepted by the automaton). The function simply enumerates all the
outgoing arcs from the current state and returns the corresponding labels in a list.
This function is called before the token prediction model (RNN + softmax), so that its result
can be used to restrict the application of softmax; the actual integration model
is discussed in detail in subsection 4.2. 


\textbf{\textit{passToken}}: For any model that predicts the output in an incremental and sequential  manner
(e.g., RNN), we want to compute the DFA state corresponding to a partial output in a similar
and lock-step fashion, so that computations in previous steps do not need to be repeated. We achieve this
by maintaining a global state, called \temv{current\_state}, which is the state reached after reading the prefix that
has been produced by the neural model up to this point.
To update the global state, the function \textit{passToken}
is called, which simply searches for the arc (`the'  again due to the DFA property) that has
the currently predicted token as a label, and then transitions to the next state via that arc.
Once this is done, the new global state will represent all the predictions made so far.
 
\paragraph{Time Efficiency Concerns}
The functions \textit{nextTokens} and \textit{passToken} need to be called on every step of the output's generation, 
and therefore need to be efficient, so that the reduction of prediction space for the token-prediction model
(e.g., RNN + softmax) can lead to runtime gains. In our case, \textit{nextTokens}  returns the labels of
all outgoing arcs and \textit{passToken} performs a simple label search in addition to carrying out a state transition.
All of these operations can be performed with O(1) time complexity.

\begin{figure*}
\centering
\includegraphics[scale=0.55]{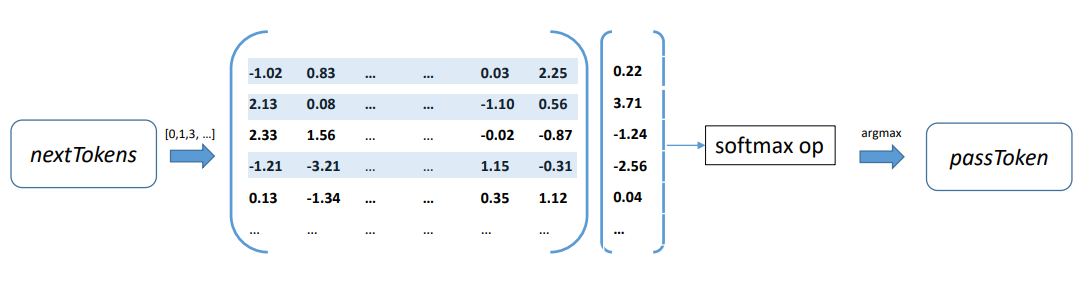}
\caption{\centering Integrating grammatical continuations into a log-linear model at one prediction step; rows selected by \textit{nextTokens} are shadowed in blue.}\label{fig:into_neural}
\end{figure*}

\subsection{Integrating Grammar into Neural Models}
After calculating the permissible next tokens, we can restrict predictions
in order to improve both prediction time and accuracy.
We apply this general strategy to the prediction layer
of our RNN-based neural network (a linear layer + softmax operation, which can be seen
as a log-linear model~\cite{DBLP:journals/corr/DymetmanX16}),
although it should be applicable to other prediction models, such as multi-class
SVMs~\cite{Duan:2005:BMS:2134810.2134843} or random forests~\cite{Ho:1995:RDF:844379.844681}. 

Figure~\ref{fig:into_neural} illustrates a concrete example of integrating the
grammar (represented as an automaton in our case) into the token prediction
model at a particular prediction step. We focus on the prediction layer of
the model, which consists of one linear layer followed by the softmax operation.
The linear layer involves a matrix of size $\card{V} \times d$, where $V$ is the LF
vocabulary and $d$ is the dimension of the vector passed from
the previous layer; the linear layer predicts scores for each of the $V$
tokens before they are passed to softmax operation.

To integrate the grammar, first, the function \textit{nextTokens} is
called to return a list of tokens allowed by the grammar at this prediction step;
the valid tokens are then translated into a list of indices, denoted by $l_c$, 
which is passed to the log-linear model. Supposing there are $k$ indices 
in the list $l_c$, we can dynamically construct another matrix of size
$k \times d$ where the \nth{\scriptsize i} row in the new matrix corresponds
to the \nth{\scriptsize j} row in the original matrix, for $j = l_c[i]$.
Figure~\ref{fig:into_neural} illustrates this process of 
choosing rows from the original matrix to construct the new matrix. 

Then the new matrix-vector product will result in  scores only for
those $k$ LF tokens that are permissible, and will then be passed to the softmax operation.
The decision function (e.g., argmax in Figure~\ref{fig:into_neural}) will
then be applied based on the softmax score, whose results will finally 
be passed to \textit{passToken} function to update the current DFA state.

\subsubsection*{Time Efficiency Concerns}

We implement \textit{nextTokens} to directly return a list of indices to avoid the cost of converting tokens to indices.  
We implemented our token prediction model in PyTorch, which supports slicing
operations so that our on-the-fly matrix construction does not need to copy
the original matrix data, but can instead just point to it. However, we find
in our experiments that even matrix construction using slicing tends to
be costly (see section 5). 

To overcome this, we observe that we can enumerate the lists returned by \textit{nextTokens}
for every DFA state, and then cache the corresponding matrices. For example, consider RCM (the Relational
Constraint Machine) in~\ref{fig:automata}. We can cache the value of \textit{nextTokens} for state 1
by precomputing  the matrix corresponding to all the enum/num fields. Doing this caching
for every DFA state can be expensive in memory; in practice, one may consider tradeoffs between
memory consumption and prediction time. 


\section{Model and Experiments}
\subsection{EQS Dataset}

Our experiments are conducted on the EQS dataset. The dataset consists of queries
paired with their LFs, which were obtained in a semi-automated manner.
The dataset contains 1981 (NL, LF) pairs as training data and 331 (NL, LF) pairs as test data.
The LF vocabulary size is 56209, most of which  consists of enum field names and values.
All the LFs can be accepted by the FSA discussed in Section \ref{sec::aut}. 

The dataset is too small to effectively learn a model that can reliably predict rare
fields or values. However, as most of the queries involve only common fields and
entities, we find in our experiments that our neural semantic parser is able to parse
a large number of those queries correctly; orthogonal research is being conducted on how
to handle more rare fields or entities.  

\subsection{Baseline Neural Model}
We use a seq2seq neural architecture as our baseline. For our encoder, we initialize the word embeddings
using Glove vectors~\cite{Pennington14glove:global}; then a Bi-LSTM is run over the question where the
last output is used to represent the meaning of the question. 
For the decoder, we again use an LSTM that runs over LF prefixes, where the LF token embeddings
are learned during training. Our decoder is equipped with an attention mechanism~\cite{DBLP:conf/emnlp/LuongPM15} 
used to attend over the output of the Bi-LSTM encoder. We use greedy decoding to predict the LFs.

We choose hyperparameters based on our previous experience with this dataset. The word and LF token
embeddings have 150 dimensions. The Bi-LSTM encoder is of dimension 150 for its hidden vector in 
each direction, therefore the decoding LSTM is of dimension 300 for its hidden vector.
We train the model with RMSprop~\cite{Tieleman2012} for 50 epochs. 

Our baseline neural model achieves 80.33\% accuracy on the test set.
Most of the errors made by our model are due to unseen fields or values;
we observe that our model also fails on queries involving compositionality
patterns that have not been seen in training, a problem similar
to those reported by~\cite{DBLP:conf/icml/LakeB18}.  

\subsection{Experimental Setups}
All our experiments were conducted on a server
with 40 Intel Xeon@3.00GHz CPUs and 380 GB of RAM.
We monitor the server state closely while conducting the experiments.

Our models are implemented in PyTorch~\cite{paszke2017automatic}, which is able
to exploit the server's multi-core architecture. The peak usage for both CPU load and
memory consumption for all our models is far below the server's capacity. 

We run all the models over the entire test dataset (331 sentences) and report the average
prediction time for each sentence. For each model, we conduct 5 such runs to calculate
the standard deviations of different runs. The standard deviations
are small in absolute and relative value.


\subsection{Results}
Integrating the LF grammar into prediction at decoding time eliminates all grammatical
errors and can therefore improve accuracy. This has been shown, for
example, by~\newcite{DBLP:conf/acl/XiaoDG16, DBLP:journals/corr/abs-1810-02720}, and indeed we obtain
similar accuracy improvements. By incorporating the grammar at decoding time at all decoding steps (using its superset represented as an automaton),
our parser is able to eliminate some grammatical errors, achieving 80.67\% accuracy on the test set, which improves our baseline model by 0.30\%. 

Table~\ref{tab:results} shows the main results of our experiments.
Our baseline neural semantic parser (NSP) takes on average 0.260 seconds to predict
the LF for a given query.  When we use the model that integrates the LF
grammar but constructs the reduced matrices on the fly (GSP-G), we find that despite the reduction of average permissible tokens (from 56209 to 6336),
the prediction time actually increases drastically to 4.416 seconds. 

To shed some light on this, we integrate the grammatically permissible next tokens
only when their number is (a) less than 500 and (b) less than $10^4$.
We observe that when the number of permissible next tokens
is small, as in case (a), integrating the grammar can indeed reduce prediction time,
indicating that the slowing is due to the dynamic matrix construction that uses the
PyTorch slicing operation, as \textit{nextTokens} and \textit{passToken} are 
called at every prediction step in all cases. 

To avoid this, we cache the reduced matrices (subsection 4.2, NSP-GC in Table~\ref{tab:results}) 
and observe that prediction time decreases in this case when more grammar integration is applied.
The best prediction time (0.067 second per query) is achieved by NSP-GC when the grammar is used
at every step. But similar speed-ups can be achieved when we are using cached matrices only for states
with a small \textit{nextTokens} set.
\begin{table}
\begin{tabular}{|c|c|c|}
\hline
Model & Avg time & Avg tokens\\
\hline
NSP &0.260 $ \pm 0.002$ & $56209$\\ 
NSP-G($500$) & $0.079 \pm 0.000$ & $9643$\\
NSP-G($10^4$) & $0.252 \pm 0.000$ & $6981$\\
NSP-G(all) & $4.416 \pm 0.029$ & $\mathbf{6336}$\\  
NSP-GC($500$) & $0.074 \pm 0.000$ & $9643$\\
NSP-GC($10^4$) & $0.069 \pm 0.000$ & $6981$\\
NSP-GC(all) & {$\mathbf{0.067 \pm 0.000}$} & $\mathbf{6336}$\\
\hline
\end{tabular}
\caption{Prediction time (in seconds) and number of permissible tokens per query on average, for our baseline neural semantic parser (NSP)
  and various models using grammar integration with caching (NSP-GC) or without (NSP-G).}
\label{tab:results}
\end{table}

\section{Related Work}

Speeding up neural models that have a softmax bottleneck is an ongoing research problem in NLP.
In machine translation, some approaches tackle the problem by moving from the prediction of word-level
units to sub-word units \cite{SennrichHB16} or characters \cite{ChungCB16}. This approach can
reduce the dimensionality of the softmax significantly, at the price of increasing the number
of output steps and thus requiring the model to learn more long-distance dependencies between
its outputs. The technique could easily be combined with the one described here; the only adaptation 
required would be to change the grammar so that it uses smaller units to define its language.
In a finite-state context, this would mean replacing transitions corresponding to a single
LF token with a sequence of transitions that construct the token from characters.
This creates potential for memory savings as well, if states in these expanded
transitions can be shared in a trie structure.    

Another approach for ameliorating a softmax bottleneck is the use of a hierarchical
softmax \cite{MorinB05}, which is based on organizing all possible output values into
a hierarchy or tree of clusters. A token to be emitted is chosen by starting at the root
cluster and then picking a child cluster until a leaf is reached. A token in this leaf
cluster is then selected. Our approach could be combined with the hierarchical softmax
method by creating a specific version of the cluster hierarchy to be associated with
every state. We would filter all impossible tokens for a state from the leaf clusters
and then prune away empty clusters in a bottom-up fashion to obtain a specific cluster.

While they have not been used in order to speed up predictions, grammars describing
possible output structures have been combined with neural models in a number of recent
papers on semantic parsing \cite{YinN17, DBLP:journals/corr/abs-1810-02720,KrishnamurthyDG17,DBLP:conf/acl/XiaoDG16,XiaoDG17}.
These papers use grammars to guide the training of the neural network model and to restrict
the decisions the model can make at training and prediction time in order to obtain
more accurate results with less data. Our approach is focused on speed improvements and
does not require any changes to the underlying model or training protocols.

Like our approach, the one presented by \newcite{DBLP:journals/corr/LHostisGA16} for machine translation tries to limit the decoding vocabulary. Their approach relies on limiting the tokens allowed during decoding to those that co-occurred frequently with the tokens in the input. Because this might rule out tokens that are needed to construct the correct output, this may decrease model performance. Our approach is guaranteed to never rule out correct outputs. For additional performance gains it should be possible to combine both approaches.

\section{Future Work}
We have used superset approximations based on finite-state automata
instead of directly using the grammar of the LF language,
which will usually be context-free. This choice is driven by the need
for an efficient implementation of \textit{passToken} and \textit{nextTokens}, 
which could be expensive for longer sequences when using a general
context-free grammar. However, for those context-free grammars that are LR \cite{Knuth65}, 
recognition can be performed in linear time, and it is easy to see that
both \textit{passToken} and \textit{nextTokens} can then be implemented with O(1)
time complexity on average. Furthermore, the caching mechanism we have proposed
for \textit{nextTokens} in this work is applicable in the case of LR grammars.
Therefore, it would be possible to implement the methods proposed here for any LR grammar,
and such grammars cover most LF languages in practical
use.\footnote{The reason being that most LF languages are designed to be machine-readable
and akin to programming languages, so they tend to be unambiguous (\egnsp, they
are fully parenthesized) and readily parsable.}

For most LF languages there will be restrictions on the logical types of
expressions that can occur in certain positions. We can detect some
of these restrictions in our finite-state automata, but in general a type
system could capture well-formedness conditions that cannot be easily
expressed with FSAs, or even in context-free grammars. It would be interesting
to investigate how more expressive type checking can be integrated
into our present framework in a more general setting.

\section{Conclusion}

We propose a method to improve the time efficiency of seq2seq models
for semantic parsing using a large vocabulary. We show that one can leverage
a finite-state approximation to the LF language in order to speed up neural
parsing significantly. Given a context-free grammar for the LF language, our strategy
is general and can be applied to any model that predicts the output in a sequential manner.

In the future we will explore alternatives to finite-state automata, which potentially
characterize the relevant LF languages exactly while still allowing for efficient 
computation of admissible next tokens. We also plan to experiment with additional datasets.

\section*{Acknowledgments} We would like to thank Mohamed Yahya for a number of insightful
comments and suggestions.

\newpage
\bibliography{LF_grammar}
\bibliographystyle{acl_natbib}
\end{document}